\pdfoutput=1

\documentclass[11pt]{article}

\usepackage[review]{acl}
\usepackage{times}
\usepackage{latexsym}

\usepackage[T1]{fontenc}

\usepackage[utf8]{inputenc}

\usepackage{microtype}

\usepackage{inconsolata}

\usepackage{graphicx}

\usepackage[utf8]{inputenc}
\DeclareUnicodeCharacter{FF0C}{,}

\title{PaperHelper: Knowledge-Based LLM QA Paper Reading Assistant}

\author{\textbf{Congrui Yin\textsuperscript{1}},
 \textbf{Evan Wei\textsuperscript{1}},
 \textbf{Zhongxing Zhang\textsuperscript{1}},
 \textbf{Zaifu Zhan\textsuperscript{2}}
\\
 \textsuperscript{1}Department of Computer Science and Engineering, University of Minnesota, Twin Cities\\
 \textsuperscript{2}Department of Electrical and Computer Engineering, University of Minnesota, Twin Cities
\\
 \small{
   \textbf{Emails:} {\{yin00486, wei00329, zhan8889, zhan8023\}@umn.edu}
 }
}

\begin{document}
\maketitle
\begin{abstract}
In the paper, we introduce a paper reading assistant, PaperHelper, a potent tool designed to enhance the capabilities of researchers in efficiently browsing and understanding scientific literature. Utilizing the Retrieval-Augmented Generation (RAG) framework, PaperHelper effectively minimizes hallucinations commonly encountered in large language models (LLMs), optimizing the extraction of accurate, high-quality knowledge. The implementation of advanced technologies such as RAFT and RAG Fusion significantly boosts the performance, accuracy, and reliability of the LLMs-based literature review process. Additionally, PaperHelper features a user-friendly interface that facilitates the batch downloading of documents and uses the Mermaid format to illustrate structural relationships between documents. Experimental results demonstrate that PaperHelper, based on a fine-tuned GPT-4 API, achieves an F1 Score of 60.04, with a latency of only 5.8 seconds, outperforming the basic RAG model by 7\% in F1 Score.
\end{abstract}

\section{Introduction}

Researchers lack a robust tool to aid in literature reading, given the overwhelming volume and diverse nature of academic literature. This challenge is compounded by time constraints and difficulties in accessing relevant material. A comprehensive tool integrating natural language processing (NLP) and information retrieval technologies is needed to offer personalized literature recommendations, summarizing, and key information extraction, facilitating an efficient and comprehensive understanding of the latest research developments.

\begin{figure}[t]
  \centering
  \includegraphics[width=0.50\textwidth]{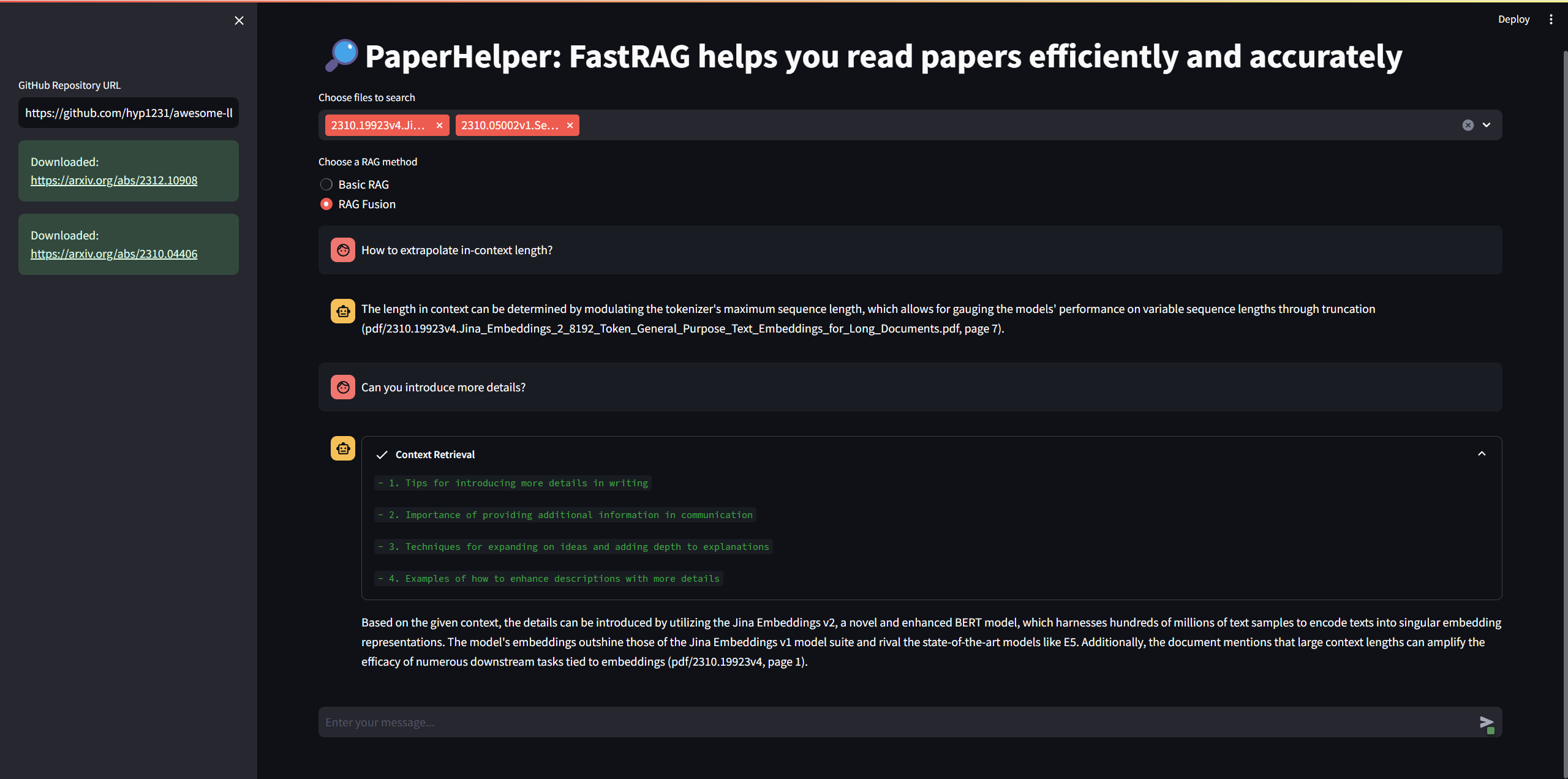}
  \caption{\textbf{PaperHelper}: A Knowledge-Based LLM QA Paper Reading Assistant}
  \label{fig1}
  \vspace{-0.55cm}
\end{figure}

Large language models (LLMs) have shown large potential in paper reading \cite{petroni2019language}. 
However, hallucination is common in LLMs， as they do not have external memory, which leads to limitations: they cannot learn new information, explain their answers, or avoid making mistakes \cite{marcus2020next}. Hybrid models combine these language models with a kind of memory that can be updated and checked \cite{karpukhin2020dense}. For example, models, such as REALM \cite{guu2020retrieval}, ORQA \cite{lee2019latent}, and Retrieval-Augmented Generation (RAG) \cite{lewis2020retrieval}, have shown promise in answering questions (QA) from a wide range of topics.

Therefore, in this project, based on RAG framework, to fully utilize the power of LLMs and minimize hallucination, we want to develop a tool called \textbf{PaperHelper} that can help researchers to efficiently review scientific literature, as illustrated in Figure \ref{fig1}. The main contributions are summarized as follow
\begin{enumerate}
    \item We design an end-to-end pipeline deployed using Streamlit, which provides a user-friendly interface and supports batch importing.
    \item We implement the basic RAG and RAG Fusion frameworks in PaperHelper, with the RAFT framework to effectively reduce the hallucinations and enhance retrieval relevance.
    \item We do parallel generating, where generative tasks would be applied on references based on relevance ranking.
    \item We integrate certain structural relationships with the extracted knowledge.
\end{enumerate}

\section{Related Work}
RAG \cite{gao2023retrieval} techniques explore the integration of external knowledge base with the LLMs to address the limitations in LLMs, such as hallucination and nontransparent reasoning process. By leveraging external knowledge databases, RAG can enhance the accuracy and transparency of LLMs. Our research focuses on the potential of RAG to significantly improve the performance, reliability, transparency, and interactivity of LLMs, where RAG could be applied in a wide range of AI tasks, e.g., paper reading, paper writing, text extraction, etc.

\textbf{RAG.} In the Open-Domain Question Answering (ODQA) domain, Sun et al. \cite{sun2024harnessing} explored the LLMs with multiple roles to enhance ODQA. However, it faces the challenges of scalability and efficiency, which would hinder its capability of leveraging the RAG framework to develop a user-friendly tool aimed at aiding researchers. Additionally, it has a significant computational burden related to LLMs. To boost the LLMs' performance in scientific research, L\'ala et al. \cite{lala2023paperqa} suggested a RAG-enhanced research agent, which relies heavily on scientific literature and struggles with computational inefficiencies, due to its reliance on modular components for iterative adjustments. Moreover, it could not simplify the user experience and limits its interactivity. Zhang et al. \cite{zhang2024raft} proposed the Retrieval Augmented Fine-Tuning (RAFT) framework, designed to improve the performance of pre-trained LLMs in domain-specific RAG settings, by teaching LLMs to effectively explore external documents in the inference process.

\textbf{Paper reading.} With the development of RAG and demands of paper-reading assistant tools, there are several tools or research work aiming at improving reading efficiency, such as ScholarPhi \cite{head2021augmenting} and Papr Readr Bot \cite{foo2022papr}. Jiang et al. \cite{jiang2024bridging} proposed a new paper interpretation system for reading and interpreting academic papers. However, it is limited by its flexibility and adaptability to the latest research findings, which could not effectively resolve the issues associated with outdated knowledge and hallucinations. Therefore, there is a lack of a more comprehensive and updated review of literature in paper reading task.
Additionally, Semantic Reader Project \cite{lo2023semantic} aims to enhance the literature reading experience with the interactive interfaces that provide relevant documents to readers, contributing to improving user experience and facilitating a deeper understanding of academic papers. CiteRead \cite{rachatasumrit2022citeread} enhanced the academic paper reading experience by integrating localized citation contexts, which help readers understand how citations are used in later research but also enriches the reader's understanding of the paper's impact and relevance. However, these tools is not flexible and do not provide a personalized reading experience or precise ranking of related references for researchers. Without the use of RAG or additional techniques, these tools could easily trapped in hallucinations and outdated knowledge, leading to imprecise retrieval of papers.

\section{Methods}
\subsection{Overview}
Our team has introduced the \textbf{PaperHelper} QA Assistant, dedicated to leveraging the RAG framework to develop a tool that aids researchers in efficiently reviewing scientific literature. The objective is to maximize the advantages of LLMs while minimizing instances of hallucination, thereby swiftly extracting high-quality, accurate knowledge to the greatest extent possible. Moreover, we aim to assist researchers by providing personalized literature recommendations, summarizations, and information extraction.

\begin{figure*}[htpb]
    \centering
    \includegraphics[width=1.0\textwidth]{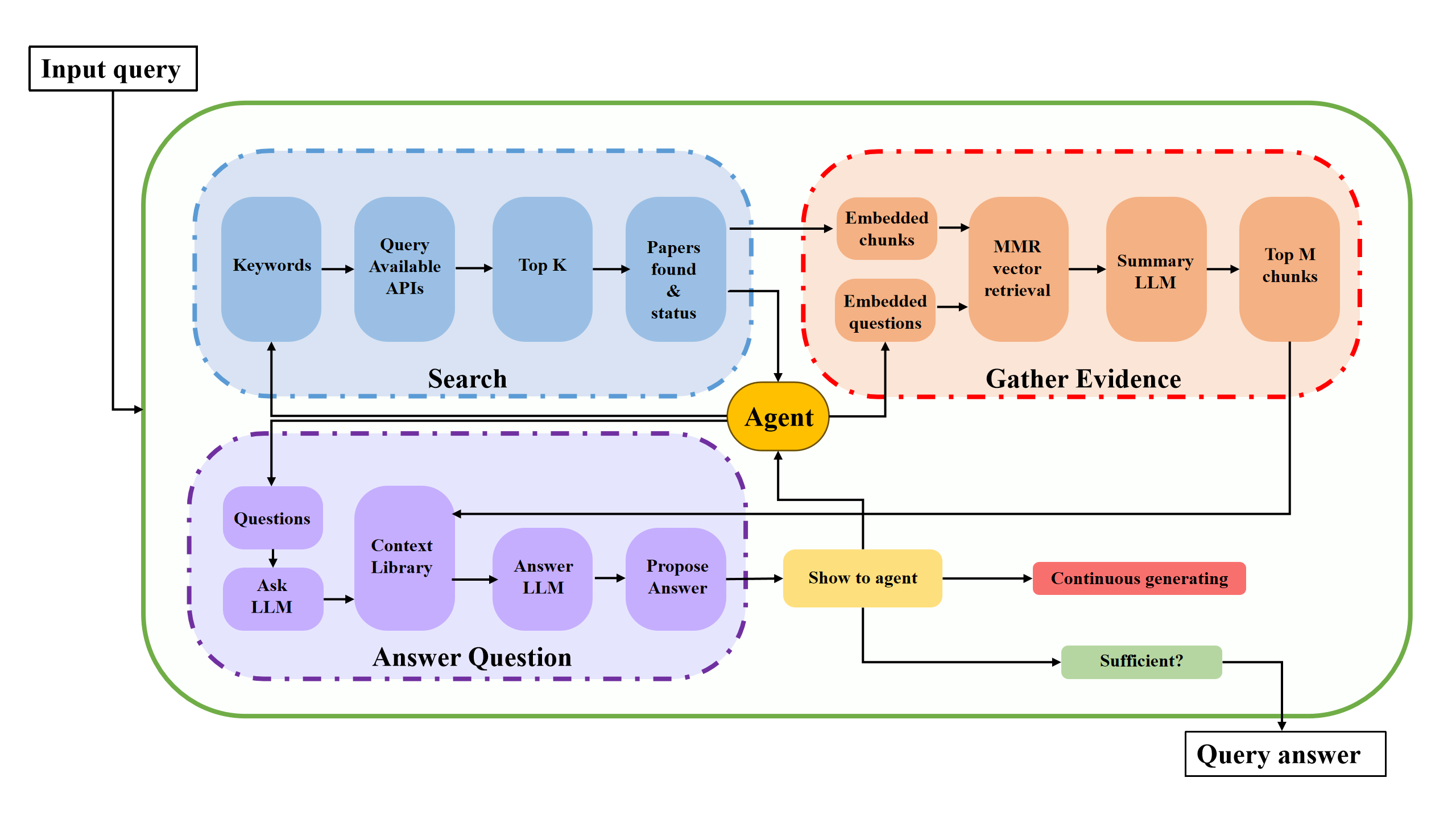}
    \caption{Schematic diagram of \textbf{PaperHelper}}
    \label{fig:enter-label}
\end{figure*}

As Figure \ref{fig:enter-label} shows, the assistant utilizes three tools: search, gather evidence, and answer questions. These tools enable it to find and parse relevant full-text research papers, identify specific sections in the paper that help answer the question, summarize those sections with the context of the question (called evidence), and then generate an answer based on the evidence. It is an agent so that the LLMs orchestrating the tools can adjust the input to paper searches, gather evidence with different phrases, and assess if an answer is complete. 

We create an end-to-end pipeline, deployed using Streamlit, which provides a user-friendly interface. In addition, it supports batch importing via URL link.
\subsection{Implementation of RAG Fusion}
\begin{figure}[t]
  \centering
  \includegraphics[width=0.50\textwidth]{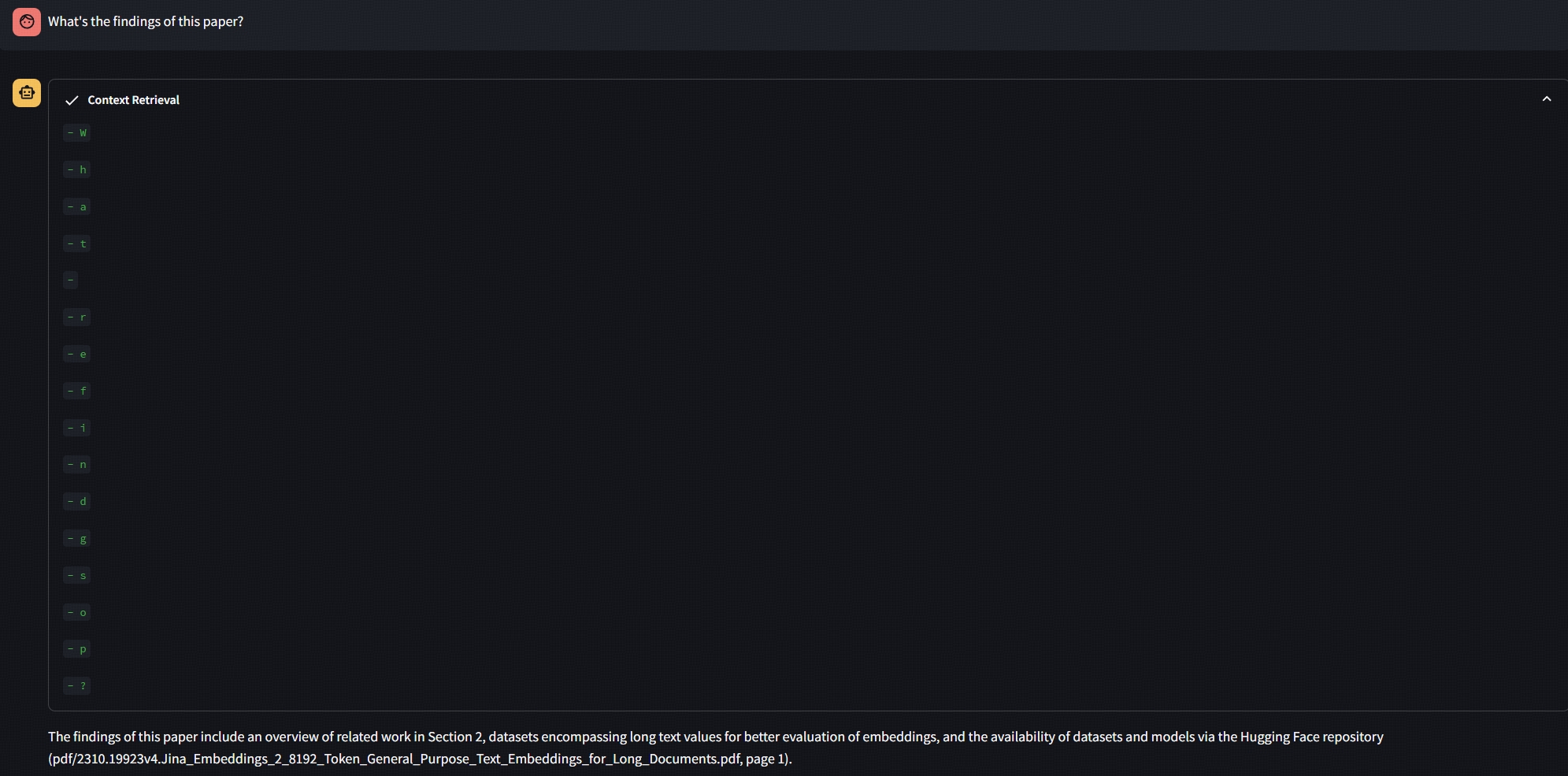}
  \caption{\textbf{RAG}: The basic RAG simply splits the search prompt into simple words in a crude manner, and may produce certain spelling illusions without truly understanding the user's intent.}
  \label{fig3}
  \vspace{-0.25cm}
\end{figure}

\begin{figure}[!ht]
  \centering
  \includegraphics[width=0.50\textwidth]{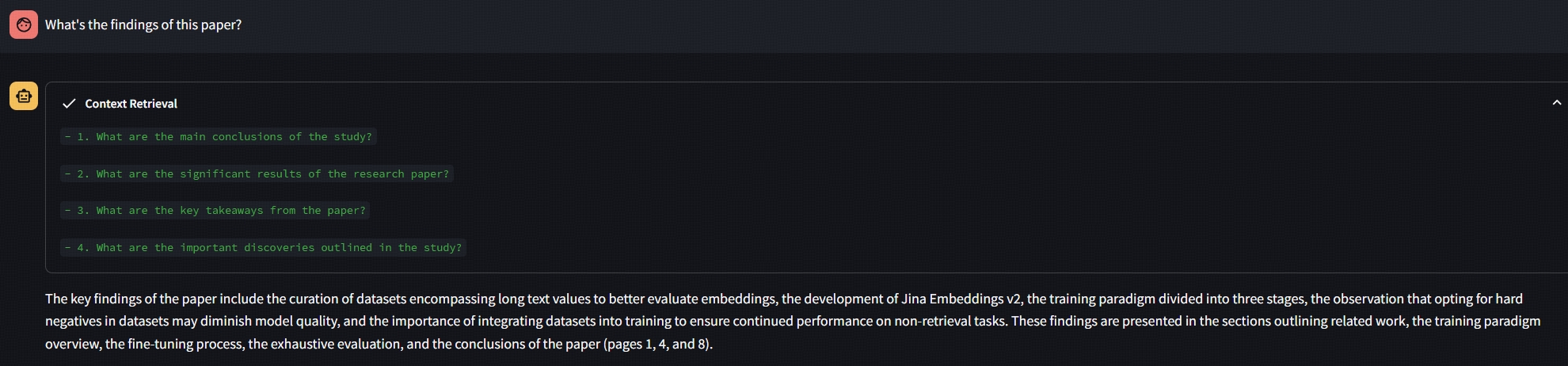}
  \caption{\textbf{RAG Fusion}: The design concept of RAG Fusion encompasses auto-prompting capabilities, addressing the common issue where users may struggle to articulate their search queries. RAG Fusion systematically captures multiple dimensions of the user's information needs, thereby delivering a comprehensive output that is enriched with an understanding of the user's intent.}
  \label{fig4}
  \vspace{-0.55cm}
\end{figure}

Compared to basic RAG shown in Figure \ref{fig3}, our system has integrated the RAG Fusion \cite{Rackauckas2024} method, RAG Fusion integrates the capabilities of RAG with Reverse Rank Fusion through the generation of multiple queries, which are subsequently reordered based on reciprocal scoring. This methodology facilitates the fusion of both documents and their corresponding scores. The effectiveness of RAG Fusion is assessed through manual evaluation, focusing on the accuracy, relevance, and comprehensiveness of the responses provided. From Figure \ref{fig4}, it can be seen that this approach enables RAG Fusion to deliver precise and comprehensive answers by contextualizing the original query from diverse perspectives.

\subsection{Implementation of RAFT}
Our system also has integrated the RAFT \cite{zhang2024raft} method. This approach enhances the capability of LLMs in specific RAG tasks by leveraging the core idea that if LLMs can "learn" documents in advance, it can improve RAG's performance. 

We finetuned the OpenAI API using 5,000 domain-specific papers from the field of machine learning to augment the knowledge of \textbf{PaperHelper} within the machine learning domain, thereby assisting machine learning scientists in reading papers more efficiently and accurately.

This is an example of the effective implementation of RAFT, when we use \textbf{PaperHelper} to read the paper "Attention is All You Need" (2017, before RAG was implemented and LLMs were proposed):

\noindent\textbf{Input: }What is RAG?

\noindent\textbf{RAG Fusion (without fine-tuning): }RAG (Region Adjacency Graph): In image processing, this is a graphical data structure used to represent the relationship between neighboring regions in an image, often used in image segmentation.

\noindent\textbf{RAG Fusion (with RAFT): }RAG (Retrieval-Augmented Generation): In the field of Large Language Model, RAG is a technique that combines information retrieval and generative modeling to improve the quality and accuracy of responses.

We can find that through the RAFT method, the model integrates cutting-edge knowledge, enabling readers to further explore academic papers based on current information rather than providing outdated and misleading content.

\begin{figure}[t]
  \centering
  \includegraphics[width=0.50\textwidth]{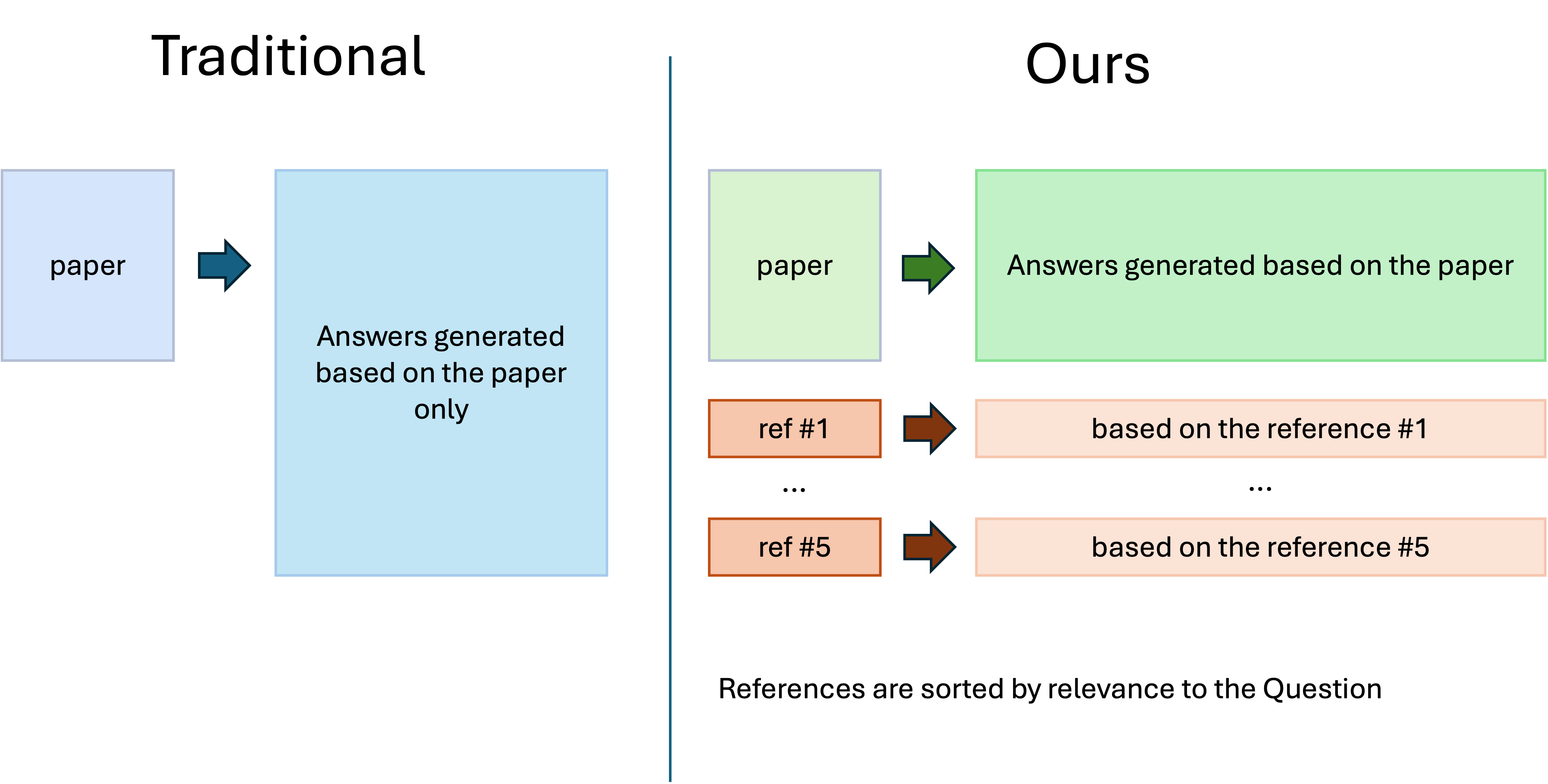}
  \caption{\textbf{Parallel Generating}: Generative tasks could also be applied to references based on relevance ranking.}
  \label{fig5}
  \vspace{-0.25cm}
\end{figure}

\begin{figure}[t]
  \centering
  \includegraphics[width=0.50\textwidth]{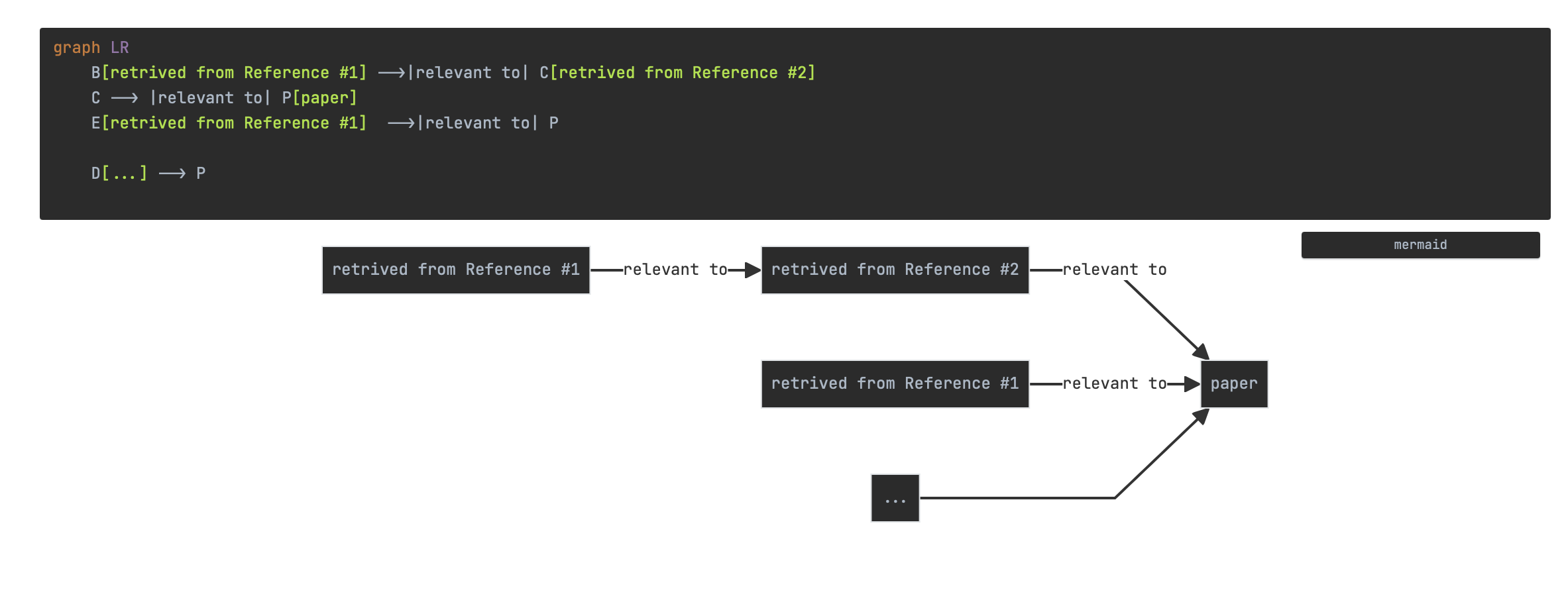}
  \caption{Show the relationship between reference papers.}
  \label{fig6}
  \vspace{-0.55cm}
\end{figure}

\subsection{Reference Knowledge Graph}

With the implementation of RAFT, we can extract the reference section at the end of articles more efficiently. First, we use RAG to traverse all the references in the article. Then, based on the knowledge from the LLMs, we refine the information using the top-k algorithm to identify the literature most relevant to the article, as shown in Figure \ref{fig5}.

As Figure \ref{fig6} illustrates, we can use the Mermaid format in Markdown syntax to represent the relationships between articles. The LLMs, based on the selected articles, will read the content using RAG and extract relationships based on the queries, generating Mermaid code, which only requires rendering to visually display the connections between the cited papers.

We can find that through the RAFT method, the model integrates cutting-edge knowledge, enabling readers to further explore academic papers based on current information rather than providing outdated and misleading content.

For example, When adding a reference, it can provide more accurate responses than GPT-4:

\noindent\textbf{Input: }According to this paper, How to expand the context length?

\noindent\textbf{Paper Helper: }The context length can be expanded by combining the two papers through a multi-stage contrastive training, utilizing a vast collection of weak pairs for domain generalization and supervised finetuning with annotated text triples to enhance the model's performance across various tasks (pdf/2310.19923v4.Jina\_Embeddings\_2\_8192, page 1). Additionally, the observation that longer sequence lengths may not always yield better outcomes due to the initial paragraph of a Wikipedia article typically mentioning the country the city is in suggests that an increase in sequence length doesn't always result in improved performance (pdf/2310.19923v4.Jina\_Embeddings\_2\_8192, page 8).

\noindent\textbf{GPT-4: }Expanding the context length in a text embedding model involves adjusting parameters or hyper-parameters related to the model architecture or the input data. Here are some general strategies you can consider:

\begin{itemize}
    \item Model Architecture: Some deep learning models, such as Transformers, allow for longer context lengths by adjusting the number of layers or the attention mechanism. Increasing the number of layers can potentially capture more contextual information, but it also increases the computational cost.
    \item Model Architecture: Some deep learning models, such as Transformers, allow for longer context lengths by adjusting the number of layers or the attention mechanism. Increasing the number of layers can potentially capture more contextual information, but it also increases the computational cost.
    \item Hardware Acceleration: Using specialized hardware such as GPUs or TPUs can enable you to handle larger context lengths by speeding up computation and memory access.
\end{itemize}

\section{Experimental Results}
We evaluate our system performance on 100 papers (test set) from MLArxivPapers \cite{arxiv_org_submitters_2024}, setting $k = 10$ (top-$k$), generating $500$ tokens, and using GPT4-32k (as Table \ref{Table1}) shows and Llama3-8b (as Table \ref{Table2}  shows). 

In our study, we employ two key metrics: precision and recall. \textbf{Precision quantifies the proportion of retrieved results that are relevant, while recall assesses the percentage of all relevant documents that the retrieval system successfully identifies.} These metrics are integrated to calculate the F1 Score, which serves as a harmonic mean of precision and recall, providing a balanced measure of the retrieval system's accuracy and completeness. 

$$
F1 \ Score= \frac{2 \cdot (Precision \cdot  Recall)}{Precision + Recall}
$$

\begin{table}[ht]
\centering
\caption{\textbf{The Performance of GPT4 API}}
\label{Table1}
\begin{tabular}{lcc}
\hline
\textbf{Methods}        & \textbf{F1  (\%)} & \textbf{Latency (s)}  \\
\hline
RAG              & 53.14            & 5.7 \\
RAG Fusion           & 58.16            & 5.6 \\
RAG Fusion + RAFT              & 60.04            & 5.8 \\
\hline
\end{tabular}
\end{table}

\begin{table}[ht]
\centering
\caption{\textbf{The Performance of Llama3-8b}}
\label{Table2}
\begin{tabular}{lcc}
\hline
\textbf{Methods}        & \textbf{F1 (\%)} & \textbf{Latency (s)}  \\
\hline
RAG              & 46.89            & 14.6 \\
RAG Fusion           & 47.81            & 13.9 \\
RAG Fusion + RAFT              & 49.44            & 14.7 \\
\hline
\end{tabular}
\end{table}

From the quantitative comparison, we can see that with the assistance of RAG, the F1 score is higher than the baseline model, and the RAG Fusion + RAFT model achieves the best F1 score, showcasing exploiting multiple knowledge extraction could effectively enhance the LLMs’ retrieval capacity. 

We also test different vector databases such as Faiss (as Table \ref{Table3} shows), Milvus (as Table \ref{Table4} shows) and Qdrant (as Table \ref{Table5} shows) with GPT4 API, here are the results:

\begin{table}[ht]
\centering
\caption{\textbf{The Performance of Faiss Vector Database}}
\label{Table3}
\begin{tabular}{lcc}
\hline
\textbf{Methods}        & \textbf{F1 (\%)} & \textbf{Latency (s)}  \\
\hline
RAG              & 53.14            & 5.7 \\
RAG Fusion           & 58.16            & 5.6 \\
RAG Fusion + RAFT              & 60.04            & 5.8 \\
\hline
\end{tabular}
\end{table}

We have found that the choice of vector database does not fundamentally affect the outcomes, except that Milvus is slightly faster due to its support for GPU acceleration. This indicates that the performance of RAG largely depends on the model, the data itself, and the data distribution strategy. Furthermore, we provide a more detailed discussion of the limitations and failure cases in \textbf{5.4 Limitations}.

\begin{table}[ht]
\centering
\caption{\textbf{The Performance of Milvus Vector DataBase}}
\label{Table4}
\begin{tabular}{lcc}
\hline
\textbf{Methods}        & \textbf{F1 (\%)} & \textbf{Latency (s)}  \\
\hline
RAG              & 53.15           & 5.5 \\
RAG Fusion           & 58.17           & 5.4 \\
RAG Fusion + RAFT              & 60.03            & 5.6 \\
\hline
\end{tabular}
\end{table}

\begin{table}[ht]
\centering
\caption{\textbf{Performance of Qdrant Vector Database}}
\label{Table5}
\begin{tabular}{lcc}
\hline
\textbf{Methods}        & \textbf{F1 (\%)} & \textbf{Latency (s)}  \\
\hline
RAG              & 53.12            & 5.8 \\
RAG Fusion           & 58.13            & 5.7 \\
RAG Fusion + RAFT              & 60.05            & 5.8 \\
\hline
\end{tabular}
\end{table}

\section{Discussion}
Since RAG started getting hot last year, task generalization in the field has advanced significantly. Enhancing our comprehension of prompts and instructions, along with pinpointing failure patterns, will enable us to develop more robust and resilient NLP systems for future applications.
\subsection{Replicability}
Our results were obtained by using the OpenAI API and Meta-Llama-3-8B-Instruct model, which is available on Huggingface. These models can be easily accessed by others as well. In addition, each of our modules is well encapsulated, and deployed via Streamlit, and a detailed README document was written to teach people how to deploy it, allowing for easy reproduction.
\subsection{Datasets}
We use a partial of the \textbf{MLArxivPapers} dataset (Due to limitations in resources, including but not limited to money, computational power, and time, we selected only about 5,000 papers) to finetune our model, which is an unlabelled collection of over 104K papers related to machine learning and published on arXiv.org between 2007-2020. The dataset includes around 94K papers (for which LaTeX source code is available) in a structured form in which each paper is split into a title, abstract, sections, paragraphs and references. Additionally, the dataset contains over 277K tables extracted from the LaTeX papers. Compared to other commonly used datasets like arXivQA or arXivRaw, the uniqueness of our dataset lies in its exclusive selection of machine learning papers from the field of computer science. This dataset provides machine learning researchers with a better platform to understand domain-specific knowledge, enhancing the generalizability across the field. It allows for the training and testing of NLP models on the unique categorizations of arXiv papers.
\subsection{Ethics}
Our project aims to explore whether LLMs can assist computer scientists in becoming more efficient at reading academic papers, and has no direct harm or risk to society. While LLMs have the potential to reduce reading time, thereby enhancing research efficiency, they still exhibit significant issues with hallucinations when dealing with complex questions that require human subjective intelligence. The primary limitations of using LLMs are due to training data and the inherent structure of LLMs themselves; hallucination issues are inevitable and can lead to erroneous perspectives, potentially impacting researchers' judgments. For newer domain-specific terms (such as "RAG"), if the relevant papers are not provided, it might incorrectly interpret these terms (e.g., as "Region Adjacency Graph," a graphical data structure used for representing relationships between adjacent regions in images, commonly used in image segmentation).
\subsection{Limitations}
Though \textbf{PaperHelper} is useful for machine learning scientists, the hallucinations inherent in LLMs always exist and can still lead to some misinformation. Our system has the following limitations:
\begin{itemize}
    \item \textbf{RAG relies excessively on the information in the input text, and it typically performs poorly when dealing with information not present in the text: }Consider the scenario that we are building a Harry Potter Q\&A system. We imported all Harry Potter stories into a vector database. Now, a question arises: how many heads does a dog have? Most likely, the system will answer three, because there are mentions of a huge dog that has three heads, and the system has no idea how many heads a normal dog may have.
    \item \textbf{Every process of RAG system is lossy: }The process of RAG involves chunking the text and generating embeddings for these chunks, then retrieving the chunks through semantic similarity search. Finally, it generates a response based on the text from the top-k chunks. All these processes are lossy, which means there is no guarantee that all information will be preserved in the final result.
    \item \textbf{PaperHelper is unable to recognize figures, which are often the most crucial part of an article: }Existing methods face significant challenges in converting paper images into vectors due to three primary limitations: (1) The information in current image data is dense and abstract, exceeding the interpretative capabilities of LLMs. (2) Even the most advanced multimodal models, like GPT-4V, struggle to comprehensively extract and correlate the figures, tables, and their interconnections found within articles. (3) A notable absence of datasets enriched with expert feedback hampers progress, as many paper images are not even identifiable within their own articles, and their full meaning can only be extracted by humans. Unfortunately, such enriched datasets are currently unavailable online. Consequently, while the RAG system performs well with pure text, the true value of an article often resides in its visual elements, which remain inadequately addressed.
\end{itemize}

\section{Conclusion}
In summary, our project successfully developed \textbf{PaperHelper}, a robust tool designed to enhance researchers' ability to efficiently navigate and understand scientific literature. Utilizing the RAG framework, \textbf{PaperHelper} effectively reduces the hallucinations common in LLMs and optimizes the extraction of accurate, high-quality knowledge. Our implementation of advanced techniques such as RAFT and RAG Fusion has significantly improved the performance, accuracy, and reliability of LLM-based literature review processes. We have also designed a user-friendly interface that facilitates batch downloading of literature and illustrates the relationships between documents using the Mermaid format.

Furthermore, we evaluated the system using a fine-tuned GPT4 API on a test set, primarily using the F1 Score as our evaluation metric. Experimental results show that \textbf{PaperHelper} performs better with the combination of RAG Fusion and RAFT methods compared to Basic RAG. The system's performance highly depends on the model, data, and data distribution itself, rather than on which vector database is used. Although our system has achieved significant results, there is still considerable room for improvement. Our system does not support reading Figures and cannot display content not present in the articles. However, our work still provides substantial domain-specific knowledge to researchers. With the increasing frequency of publications in machine learning today, it could benefit researchers by allowing them to quickly read and understand the connections between articles in a domain. As the models continue to advance, we anticipate that enhanced multimodal models will help us read and fully understand the content of Figures, making them more expert tools for scientific research. Additionally, we hope it provides a good experience for both novice and veteran machine learning researchers, offering answers to questions and related justifications, thus contributing to a more pleasant and suitable experience for scientific advancement worldwide.

\bibliography{main.bib}

\appendix

\label{sec:appendix}


\end{document}